\documentclass[10pt,twocolumn,letterpaper]{article}

\usepackage{soul}
\usepackage{xfrac}
\usepackage{amsfonts}
\usepackage{booktabs}
\usepackage{algorithm}
\usepackage{algorithmic}
\usepackage{bbm}
\usepackage{bm}

\usepackage{amsmath}
\usepackage{amsthm}
\usepackage{mathtools}

\newcommand{\x}{{\bf x}}

\newcommand{\y}{{\bf y}}

\newcommand{\be}{{\bf e}}

\newcommand{\bPhi}{{\bm \Phi}}
\newcommand{\bpsi}{{\bm \psi}}
\newcommand{\bPsi}{{\bm \Psi}}

\newcommand{\bW}{{\bf W}}

\newcommand{\bX}{{\bf X}}

\newcommand{\bV}{{\bf V}}

\newcommand{\bone}{\mathbbm{1}}

\newcommand{\bS}{{\bf S}}

\newcommand{\bTheta}{{\bf \Theta}}

\newcommand{\bomega}{{\bm \omega}}

\newcommand{\bphi}{{\bm \phi}}

\newcommand{\bbP}{\mathbb P}
\newcommand{\bbR}{\mathbb R}

\newcommand{\bbE}{\mathbb E}

\newcommand{\cX}{\mathcal X}

\newcommand{\cH}{\mathcal H}

\newcommand{\cF}{\mathcal F}

\newcommand{\cB}{\mathcal B}

\newcommand{\cD}{\mathcal D}
\newcommand{\cO}{\mathcal O}

\newcommand{\cE}{\mathcal E}

\newcommand{\cC}{\mathcal C}

\newcommand{\cY}{\mathcal Y}

\newcommand{\fP}{\mathfrak P}

\newcommand{\fK}{\mathfrak K}

\DeclareMathOperator*{\argmin}{arg\,min}

\newtheorem{theorem}{Theorem}

\newtheorem{lemma}{Lemma}
\newtheorem{assumption}{Assumption}

\newtheorem{definition}{Definition}

\newtheorem{remark}{Remark}

\usepackage{mathtools}

\allowdisplaybreaks
\usepackage[pagenumbers]{cvpr} 

\usepackage{amsmath}
\usepackage{amssymb}
\usepackage[table, dvipsnames]{xcolor}
\usepackage{multirow}

\usepackage[pagebackref,breaklinks,colorlinks]{hyperref}

\usepackage[capitalize]{cleveref}
\crefname{section}{Sec.}{Secs.}
\Crefname{section}{Section}{Sections}
\Crefname{table}{Table}{Tables}
\crefname{table}{Tab.}{Tabs.}

\def\cvprPaperID{11297} 
\def\confName{CVPR}
\def\confYear{2022}

\usepackage{xcolor}

\begin{document}

\title{Adaptive Methods for Aggregated Domain Generalization
}

\newcommand*\samethanks[1][\value{footnote}]{\footnotemark[#1]}

\author{Xavier Thomas\thanks{Manipal Institute of Technology}
\\ \small{xavier.thomas1@learner.manipal.edu}
\and Dhruv Mahajan\thanks{Facebook AI Research, Meta}
\\ \small{dhruvm@fb.com}
\and Alex Pentland\thanks{MIT}
\\ \small{pentland@mit.edu}
\and Abhimanyu Dubey\samethanks[2]
\\ \small{dubeya@fb.com}}
\maketitle

\begin{abstract}
   Domain generalization involves learning a classifier from a heterogeneous collection of training sources such that it generalizes to data drawn from similar unknown target domains, with applications in large-scale learning and personalized inference. In many settings, privacy concerns prohibit obtaining domain labels for the training data samples, and instead only have an {\em aggregated} collection of training points. Existing approaches that utilize domain labels to create domain-invariant feature representations are inapplicable in this setting, requiring alternative approaches to learn generalizable classifiers. In this paper, we propose a {\em domain-adaptive} approach to this problem, which operates in two steps: (a) we cluster training data within a carefully chosen feature space to create pseudo-domains, and (b) using these pseudo-domains we learn a {\em domain-adaptive} classifier that makes predictions using information about both the input and the pseudo-domain it belongs to. Our approach achieves state-of-the-art performance on a variety of domain generalization benchmarks {without using domain labels whatsoever}. Furthermore, we provide novel theoretical guarantees on domain generalization using cluster information. Our approach is amenable to ensemble-based methods and provides substantial gains even on large-scale benchmark datasets.
\end{abstract}

\section{Introduction}
\label{sec:intro}
The problem of {\em domain generalization} addresses learning a classifier from a random subset of training {\em domains}, with the objective of generalizing to unseen test domains~\cite{blanchard2011generalizing}. Research on this problem has seen an explosion in interest recently, with a majority of approaches focusing on learning {\em domain-invariant} feature representations~\cite{li2018deep, muandet2013domain, li2018domain, ghifary2015domain, arjovsky2019invariant}. The general idea behind such methods is to learn feature representations that reduce {\em domain-specific} variance, which can, under suitable assumptions, be shown to minimize generalization error (see, e.g., ~\cite{ben2010theory}).

From a practical perspective, the most relevant application of domain generalization is in large-scale learning~\cite{kairouz2019advances}, where data is gathered from multiple (often varying) sources, and we wish to learn a model that generalizes to new sources of data, e.g., learning an image classifier from data obtained from a collection of mobile devices. Each data source exhibits unique properties, and hence it is desirable to eliminate {\em domain-specific} variation (i.e., changes across users) to reduce spurious error.

In addition to this challenge of generalization,  environments with data from multiple participants additionally present a second challenge of privacy-preservation. When data is collected from various sources (e.g., users), it is desirable to eliminate sensitive information to ensure privacy of the participants.
In the domain generalization setting, a straightforward approach to achieve this is to discard domain information altogether, i.e., {\em aggregating} all points into an anonymized dataset. However, in this problem setting, existing {\em domain-invariant} approaches present an insurmountable challenge: without domain information, it is not possible to construct appropriate regularization penalties to learn invariant features. Furthermore, recent work~\cite{gulrajani2020search} suggests that domain information may not be even necessary, as simple fine-tuning matches {\em domain invariant} methods when model selection is done properly. These reasons make domain generalization in the {\em aggregate} setting (i.e., {\em without} domain labels) an interesting open problem.

In this paper, we demonstrate that when fine-tuning from pre-trained models (as is standard in computer vision), it indeed is possible to generalize to new domains without access to the domain partitions. The fundamental insight for our approach is to recover {\em latent} domains via clustering, and subsequently bootstrap from these latent domains using {\em domain-adaptive} learning~\cite{dubey2021adaptive}. Specifically, we provide an algorithm that recovers domain information in an unsupervised manner, by carefully removing {\em class-specific} noise from features. We then use this carefully selected feature space to partition inputs and learn a {\em domain-adaptive} classifier with state-of-the-art performance. Our precise contributions are summarized as follows.
\begin{itemize}
\item We extend {\em domain-adaptive} domain generalization to an algorithm that simultaneously assigns latent domain labels to an {\em aggregated} training set via unsupervised clustering, and then runs regular fine-tuning on an augmented input space to produce a classifier that {\em adapts} to the domain corresponding to any input.

\item Additionally, we extend the theory of domain generalization via kernel mean embeddings~\cite{blanchard2011generalizing,  dubey2021adaptive} to approaches that utilize an approximate clustering of the domain space, and provide novel generalization bounds under this setting, which are applicable in tasks beyond those considered within this paper.

\item On a set of standard and even large-scale (1M+ points) domain generalization benchmarks, we demonstrate that even when the training data is aggregated, it is possible to obtain competitive performance in the domain generalization task by {\em domain-adaptive} classification. 
\end{itemize}

\section{Related Work}
\label{sec:related_work}

Our work draws from several lines of research in computer vision and machine learning, as discussed below.\\

\noindent{\bf Domain Generalization}. First proposed in the work of Blanchard \etal\cite{blanchard2011generalizing}, domain generalization is a problem gaining rapid attention in the machine learning and computer vision communities. A broad category of approaches can be summarized by {\em domain-invariant} representation learning, i.e., learning representations that eliminate domain-specific variations within the dataset. This approach was first examined in the context of {\em domain adaptation} by Ben-David \etal\cite{ben2010theory}, which was used to construct a {\em domain-adversarial} neural network in the work of Ganin \etal\cite{ganin2016domain}. Building on the work of~\cite{ganin2016domain}, several algorithms have been proposed for domain generalization~\cite{sun2016deep, li2018domain, li2018deep} via adversarial feature learning. Key differences within these approaches are based on the penalty formulation used to ensure invariant feature learning. For example, Li \etal\cite{li2018deep} utilize a maximum mean discrepancy (MMD) regularization, Sun \etal\cite{sun2016deep} use a correlation alignment, and Li \etal\cite{li2018domain} propose class-conditional adversarial learning. 

In contrast to these approaches, Arjovsky \etal\cite{arjovsky2019invariant} propose invariant risk minimization (IRM), a training method that optimizes for a robust loss function in order to provably reduce out-of-distribution error. A similar robust design philosophy has been explored in the work of Sagawa~\etal\cite{sagawa2019distributionally} via distributionally robust optimization, and a straightforward but effective interpolation strategy known as MixUp~\cite{xu2020adversarial, yan2020improve, wang2020heterogeneous}. Generalization via assuming a causal structure has also been explored in~\cite{mahajan2020domain, christiansen2021causal}. While these approaches have seen improvements on domain generalization benchmarks, recent work by Gulrajani and Lopez-Paz~\cite{gulrajani2020search} suggests that improvements obtained by {\em domain-invariant} approaches are largely dependent on hyperparameter settings and the model selection technique used, as naive ERM (vanilla training using the training data) outperforms several of these approaches when initialized properly. Teney \etal\cite{teney2020unshuffling} for the task of Visual Question Answering shows that partitioning the data into well-chosen environments can lead to better generalization, by capturing patterns that are stable across environments and discarding spurious ones.

Our setting departs from the ones considered within the above line of work as we do not assume access to the domain labels, which are imperative for learning invariant features in the methods highlighted previously. Similar to~\cite{gulrajani2020search}, our core algorithm is also vanilla ERM, and our improvements arise from using a more expressive class of functions.

\noindent{\bf Kernel Mean Embeddings}. Our design philosophy utilizes {\em kernel mean embeddings}~\cite{muandet2016kernel}, the technical tool used originally by Blanchard~\etal\cite{blanchard2011generalizing} to study the domain generalization problem. Kernel mean embeddings provide a rigorous and realizable mechanism to ``project'' probability distributions on to reproducing kernel Hilbert spaces (RKHS), and have been shown to be effective in generalization across a variety of problems, including multi-task learning~\cite{deshmukh2017multi} and reinforcement learning~\cite{dubey2021provably}. We further the analysis of learning with kernel mean embeddings from that presented in~\cite{blanchard2011generalizing, blanchard2011generalizing} by providing novel generalization bounds in the approximate setting, where we learn embeddings in an unsupervised manner via bootstrapping. 

\section{Approach}
\label{sec:approach}

\begin{figure*}[htb]
\centering
\label{fig:net_architecture}
\includegraphics[width=0.9\linewidth]{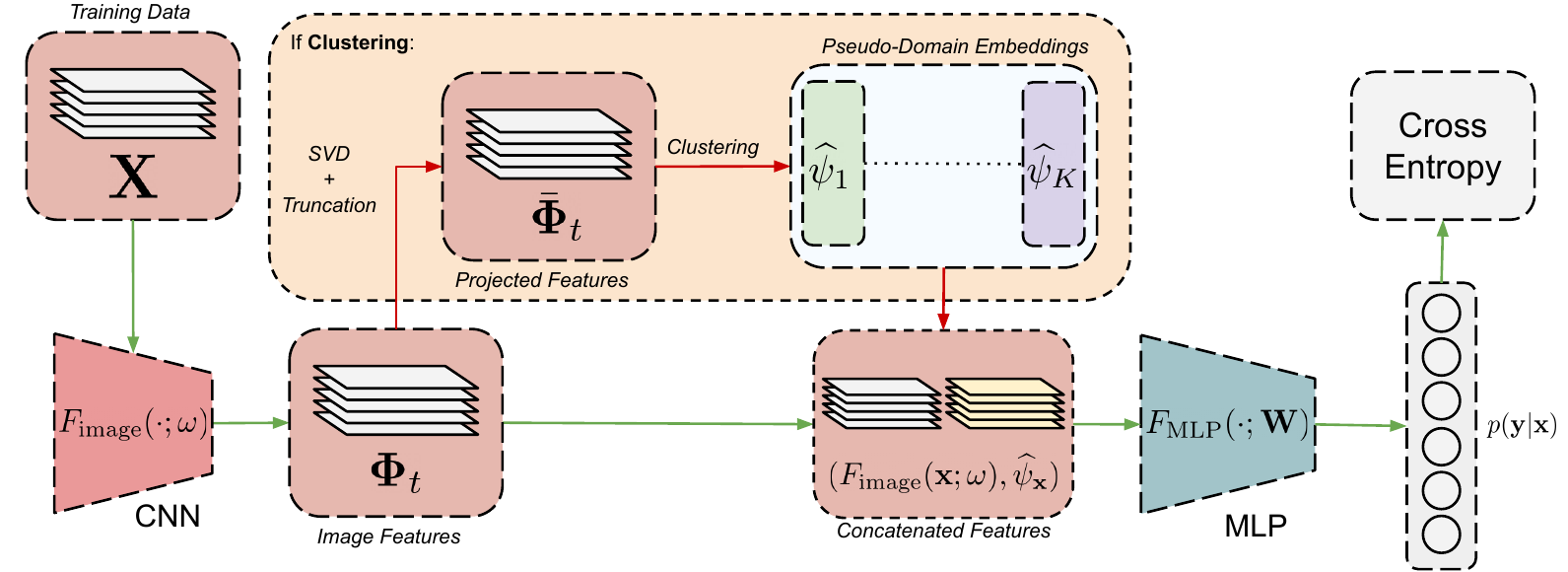}
\caption{Training Pipeline. The shaded {\color{orange} orange} area is only run when the epoch $t \in T_{\text{clust}}$; we reuse previous presudo-domain embeddings otherwise. The paths in {\color{ForestGreen} green} denote where there are gradient flows, and the paths in {\color{red} red} denote {\bf only} feed-forward operations.}
\end{figure*}

The domain generalization setting typically assumes a space of all relevant data distributions $\cD$, i.e., each domain $D \in \cD$ defines a probability distribution over the space $\cX \times \cY$.  Further, we assume that $\cD$ is endowed with a ``mother'' probability distribution $\fP$ that determines how a domain is sampled. A {\em training domain} $\widehat D(n)$ is obtained by first sampling a domain $D \in \cD$ following $\fP$ and then sampling $n$ points from $\cX \times \cY$ following $D$. The training set is constructed by sampling $N$ such domains $(\widehat D_i(n))_{i=1}^N$ and aggregating them. A test domain is constructed by sampling a domain $\widehat D_T(n_T)$ identically but discarding the labels. Since we are working in the {\em aggregated} setting, we assume that the training set only consists of a set of $n\cdot N$ points $\{\x_i, y_i\}_{i=1}^{n\cdot N}$ from all the sampled domains without the corresponding domain labels.

{\bf Domain-Adaptive Classification}. Our approach at a high level follows the {\em domain-adaptive} paradigm introduced in~\cite{dubey2021adaptive}, building on the kernel mean embedding approach for domain generalization~\cite{blanchard2011generalizing}. Consider a family of functions $\cF$. For each domain $D \in \cD$, we have that the optimal classifier within $\cF$ under a loss function $\ell$ can be given by $f_D = \argmin_{f\in\cF} \bbE_{(\x, y)\sim D}\left[\ell(f(\x), y)\right]$. Additionally, the {\em universal} optimal classifier $f_\star$ (over all domains) within the class $\cF$ can be given by $f_\star = \argmin_{f\in\cF}\bbE_{D\sim\fP}\bbE_{(\x, y)\sim D}\left[\ell(f(\x), y)\right]$. Unless we make regularity assumptions on the geometry of $\cD$, once can see that the test error of $f_\star$ on an randomly chosen domain $D' \in \cD$ can be arbitrarily worse.

The primary motivation behind {\em domain-adaptive} classification is to expand the function class to consider functions over $\cX \times \cD$, i.e., a function $F$ that takes in {\em both} the input sample $\x$ and the domain $D$ while making predictions, and attempt to ensure {\em uniformly} low error, i.e., ensure that for any arbitrarily sampled test domain $\widehat D_T$, the {\em adaptive} classifier $f_T = F(\cdot, \widehat D_T)$ incurs low risk. The central challenge in such an approach is to learn a function over each training {\em domain}. For this, we need an approach to represent each domain itself as vector, which brings us to the framework of {\em kernel mean embeddings}.

{\bf Kernel Mean Embeddings}. Kernel mean embeddings (KMEs) are a straightforward technique to compute a function over probability distributions, which also provide rigorous convergence guarantees~\cite{muandet2016kernel}. The approach outlined in~\cite{dubey2021adaptive} trains such an {\em adaptive} network by representing the domain $D$ via its kernel mean embedding $\widehat\bPsi_D$. Specifically, we can learn a {\em domain-adaptive} function $F(\x, D)$ as $F = f(\x, \widehat\bPsi_D)$, where $f$ is a neural network that takes in the {\em joint} input, i.e., the input $\x$ and the KME of the domain $D$ computed via a separate feature extractor $\bphi(\cdot)$, i.e.,  $\widehat\bPsi_D = \frac{1}{n}\cdot\sum_{i=1}^n \bphi(\x_i)$, where $\x_1, ..., \x_n$ are sampled i.i.d. from $D$. In~\cite{dubey2021adaptive}, the authors adopt this approach to learn a CNN classifier. The authors learnt the feature $\bphi$ by first training a {\em domain classifier}, and then discarding the softmax layer to use its features. The network $F$ in this case is then a CNN that first produces image features $\bphi_{\text{image}}(\x)$, which are concatenated with the domain embeddings $\widehat\bPsi_D$, followed by an MLP.



\begin{algorithm}[t!]
\footnotesize
\caption{Training Pseudocode}
\label{alg:training}
\underline{\textsc{Training}}\\
\textbf{Input.} Aggregated training data $\widehat D_{\text{tr}}$, clustering schedule $T_{\text{clust}}$\\
$K$ : \#clusters, $d_{\text{start}}$: starting component, $d_{\text{end}}$: last component.\\
\textbf{Output.} Classifier $F_{\text{image}}$ with weights $\bomega$,  $F_{\text{MLP}}$ with weights $\bW$ and clustering centroids $\{\widehat\bpsi_k\}_{k=1}^K$.

\begin{algorithmic}
\STATE \textbf{Initialize.}  Run 1 epoch of SGD using pre-trained $\bomega_0$ on training data $\widehat D_{\text{tr}}$ and obtain $\bomega_1$, discard final layer weights. Initialize new $\bW$.
\FOR{Round $t = 1$ to $T$}
\IF{$t \in T_{\text{clust}}$}
\STATE {\color{gray}// do clustering in this epoch}
\STATE $\bPhi_t \leftarrow$ {\sc Feature Extraction}($\bomega_t, \widehat D_{\text{tr}}$).
\STATE $\bar\bPhi_t \leftarrow$ {\sc SVD + Truncation}($\bPhi_t, d_{\text{start}}, d_{\text{end}}$).
\STATE $\{\widehat\bpsi_k\}_{k=1}^K, \{\widehat D_k\}_{k=1}^K \leftarrow$ {\sc Clustering}($\bar\bPhi_t, K$).
\STATE Create augmented dataset.
\ENDIF
\FOR{each batch $(\x, \widehat\bpsi_\x, y)$}
\STATE $\bPhi(\x) \leftarrow F_{\text{image}}(\x; \bomega_{\text{t}})$. {\color{gray}// compute image features}
\STATE $\hat y \leftarrow F_{\text{mlp}}(\textsc{Concat}(\bPhi(\x), \widehat\bpsi_\x); \bW_t)$. {\color{gray}// compute predictions}
\STATE $J_t \leftarrow \textsc{CrossEntropy}(\hat y, y)$. {\color{gray}// compute loss}
\STATE $\bomega_{t+1}, \bW_{t+1} \leftarrow \textsc{SGD Step}(J_t, \bomega_{\text{t}}, \bW_{t})$. {\color{gray}// gradient descent}
\ENDFOR
\ENDFOR
\end{algorithmic}
\hrulefill\\
\underline{\textsc{Inference}}\\
\textbf{Input.} Network with weights $\bomega, \bW$ and embeddings $\{\widehat\bpsi_k\}_{k=1}^K$.\\ 
\textbf{Output.} Prediction for any point $\x$.
\begin{algorithmic}
\STATE $\bPhi(\x) \leftarrow F_{\text{image}}(\x; \bomega)$
\STATE $\bphi(\x) \leftarrow \textsc{Projection}(\bPhi(\x))$ onto $[d_{\text{start}}, d_{\text{end}}]$
\STATE $\widehat\bpsi_\x \leftarrow\argmin_{k\in[K]}\lVert \widehat\bpsi_k - \bphi(\x)\rVert_2$.
\RETURN $F(\x) = F_\text{mlp}\left(\textsc{Concat}\left(\bPhi(\x), \widehat\bpsi_\x\right); \bW\right)$.
\end{algorithmic}
\label{algo:prediction}
\end{algorithm}

\subsection{Aggregated Domain Generalization} The aforementioned adaptive approach (and moreover, any approach building on using domain labels to create {\em domain-invariant} classifiers) is not feasible when the domain labels are unavailable, and we only have access to aggregated data with the knowledge that it comprises inputs from several sources. We assume that we are provided with a training set $\widehat D_{\text{tr}}$ with $T = n\cdot N$ samples ($n$ from each of the $N$ domains), but the dataset is {\em aggregated}, i.e., the domain labels are discarded. During testing, we are provided with $n_T$ samples from a fresh domain $D_T$ sampled i.i.d. from $\fP$. The objective, once again, is to construct a classifier that achieves {\em uniform} low risk on the test domains. 

To counter the lack of labels, we rely on a bootstrapping-style approach, i.e., where we cluster the training data into $K$ {\em pseudo-domains} $\{\widehat D_1, \widehat D_2, ..., \widehat D_K\}$, and then compute the kernel mean embedding of each pseudo-domain itself, and perform adaptive classification. At a high level, our training involves two steps in every epoch:

\noindent(A) {\bf Cluster Training Data}. We compute a clustering $\{\widehat D_k\}_{k=1}^K$ (also known as {\em pseudo-domains}) of the input data using an appropriately chosen feature $\bphi$. Once these clusters are obtained, we obtain the corresponding {\em pseudo} mean embeddings $\{\widehat\bpsi_k\}_{k=1}^K$ for each {\em pseudo-domain}, i.e., $\widehat\bpsi_k = (1/|\widehat D_k|)\cdot\sum_{\x\in\widehat D_k}\bphi(\x)$. After clustering, we compute the {\em augmented inputs}, i.e., where each point $\x$ in the training data is augmented with the corresponding $\widehat\bpsi_\x$ from the cluster it belongs to, to produce the augmented dataset $((\x, \widehat\bpsi_\x), y)_{\x\in \widehat D_{\text{tr}}}$. To avoid heavy computation, this clustering step is not performed every epoch but at a logarithmic schedule. Hence, if training progresses for $T$ epochs, we recompute clusters $\cO(\log(T))$ times. 

\noindent(B) {\bf ERM using augmented pseudo-embeddings}. 
We learn a function $F$ which involves a feature extractor $F_{\text{image}}$ with weights $\bomega$ (a CNN, e.g., ResNet~\cite{he2016deep}), followed by a fully-connected classifier $F_{\text{MLP}}$ with weights $\bW$. The feedforward operation for an input $(\x, \widehat\psi_\x)$ involves three steps: (a) obtaining the features $F_{\text{image}}(\x; \bomega)$; (b) concatenating the feature $F_{\text{image}}(\x)$ with the mean embedding $\widehat\bpsi_\x$ to form the augmented input $F_{\text{joint}}(\x)$; (c) computing class probabilities by feeding $F_{\text{joint}}(\x)$ through the linear layer $F_{\text{MLP}}$ with weights $\bW$, followed by a softmax operation. We use the standard cross-entropy error (ERM) to optimize the backward pass, and note that there are no gradients through $\widehat\bpsi_\x$, i.e., the centroids $\widehat\bpsi_\x$ are fixed features, treated as additional inputs (until they are recomputed). The complete algorithm is summarized in Algorithm~\ref{algo:prediction}. 

{\bf Testing}. During testing, for any input $\x_T$, we identify the nearest pseudo embedding $\widehat\bpsi_{\x_T}$ and return $F(\x_T, \bpsi_{\x_T})$. In contrast to prior work on adaptive domain generalization~\cite{blanchard2011generalizing,  dubey2021adaptive}, inference is carried out one sample at a time.

We now provide more details on how we select the feature $\bphi$ and the corresponding theoretical guarantees. 
\subsection{Obtaining Pseudo-Domains via Clustering}

{\bf Selecting $\bphi$}. Selecting an appropriate feature embedding $\bphi$ to perform clustering is imperative for our bootstrap-based approach. Our central idea is to recover {\em domain-specific} features from the existing network $F_{\text{image}}$ itself, by carefully selecting relevant directions of importance.

We first fine-tune a pre-trained network (trained on a large dataset, e.g., ILSVRC12~\cite{deng2009imagenet} as per standard practice), for 1 epoch on our target data. We then discard the final (classification) layer and set $F_{\text{image}}$ to be the remaining feature extractor. Now, we use $F_{\text{image}}$ itself to provide us with the relevant features $\bphi$. However, these features primarily contain information about the prediction problem, i.e., {\em class-specific} variance, however, we only want to extract {\em domain-specific} variance so that the resulting clustering separates the domains well.

For this, we assume that the features from $F_{\text{image}}$ can be decomposed into three broad categories: the first are {\em class-specific} features, next are {\em domain/image-specific} features, and finally, we have noise. For some parameters $d_{\text{start}}, d_{\text{end}} \leq d$, we assume that the fist $d_{\text{start}}$ principal eigenvectors capture {\em class-specific} variance and the last $d_{\text{end}}$ principal eigenvectors are primarily noise. Hence, to obtain a {\em domain-specific} clustering, we only need to consider the central $[d_{\text{start}}, d_{\text{end}}]$ portion of the spectrum. To obtain these features, in each clustering round, we first compute the image features $\bPhi = F_{\text{image}}(\bX)$ and project $\bPhi$ on to its central $[d_{\text{start}}, d_{\text{end}}]$ principal components by first performing a PCA operation (to obtain the principal components), and then truncation (to remove the irrelevant feature directions). This provides us with the resulting ``projected'' features $\bar\bPhi$. Finally, we cluster $\bar\bPhi$ into $K$ clusters ({\em pseudo-domains}), and provide the resulting centroids of each cluster $\{\widehat\bpsi_k\}_{k=1}^K$ as the KMEs for each of the $K$ pseudo-domains. 

Note that the filtering of irrelevant feature directions is a critical component of the algorithm, as demonstrated via ablation experiments in Section~\ref{sec:ablations} as well. However, we see that the algorithm is robust to the precise values of $[d_{\text{start}}, d_{\text{end}}]$ as long as they are within a reasonable range. 

{\bf A Remark on Scalability}. Centralized data operations such as clustering and finding nearest neighbors are known to be expensive and difficult to scale, however, recent work~\cite{johnson2019billion} has demonstrated that by careful quantization and efficient implementation, one can scale even to billion-scale datasets. We employ a similar approach and our experiments demonstrate that even for large-scale settings, our algorithm scales effectively.
\section{Theoretical Guarantees}
In this section we provide some theoretical guarantees for our proposed algorithm. The theoretical results are within the framework of Blanchard \etal~\cite{blanchard2011generalizing}, where our key contributions are to analysing when the true kernel mean embeddings and obtained embeddings are mismatched. At a high level, our function class $\cF$ is defined over the space $\cX \times \cD$ (i.e., the joint input-domain space), however, we {\em embed} domains into $\bbR^d$ via a {\em mean embedding} $\bPsi : \cD \rightarrow \bbR^d$, and therefore, any function $f \in \cF$ is written as $f(\x, \bPsi_D)$, where $\bPsi_D$ denotes the {\em kernel mean embedding} of $D$.

\noindent{\bf Background}. We assume a {\em compact} input space $\cX$ and assume the outputs to lie in the space $\cY = [-1, 1]$. For any Lipschitz loss function $\ell$, then the empirical loss on any domain $D$ with $n$ samples is $\frac{1}{n_i}\sum_{(\x, y) \in D} \ell(f(\x, \bPsi_D), y)$. We can define the average training error over $N$ domains with $n$ samples as,
\begin{equation*}
    \widehat{L}_N(f, \bPsi) \triangleq \frac{1}{n\cdot N}\sum_{i \in [N]}\sum_{(\x, y) \in D_i} \ell(f(\x, \bPsi_{D_i}), y).
\end{equation*}
Similarly, our benchmark is to compare the above with the {\em expected} risk, i.e., the error obtained by $f$ in the limit of infinite samples.
\begin{equation*}
    L(f, \bPsi) \triangleq \underset{D\sim\fP}{\bbE}\left[\underset{(\x, y)\sim D}{\bbE}\left[\ell(f(\x, \bPsi_D), y\right]\right].
\end{equation*}
Following~\cite{blanchard2011generalizing}, the space $\cF$ we consider is defined by product kernels, i.e., the kernel $\kappa$ can be decomposed into a product of two separate kernels $k_P$ (which depends on the domain via $\bPsi$) and $k_X$ (which depends on the inputs $\x$). We defer more kernel assumptions to the Appendix. 

\noindent{\bf Results}. Our key result is a uniform bound on the {\em excess risk} of using $K$ pseudo-domain centroids $\widehat\bPsi$ instead of the true kernel mean embeddings $\bPsi$, since computing the true kernel mean embeddings would require the domain labels. We make two assumptions on the data distribution $\fP$ and feature space $\bphi$ to obtain our generalization bound. We state the assumptions informally, and provide detailed technical explanations with examples in the Appendix.

\begin{assumption}[$d_\star$-Expressivity, Informal]
\label{ass:expressivity}
We assume that the feature space $\bphi$ is expressive with a parameter $d_\star \ll d$ for the distribution $\fP$, i.e., it requires on average $d_\star$ dimensions to cover the domain space $\cD$. Specifically, if we assume that there exists an optimal clustering (with infinite samples) into $K$ partitions within $\fP$ whose centroids are given by $\bPsi_\star$, then, we assume that for any domain $D\in\cD$, $\min_k\lVert \bPsi_D - \bPsi_{\star, [k]}\rVert_2 \leq \cO\left(\frac{1}{K^{d_\star}}\right)$.
\end{assumption}

The above assumption implies that $\bphi$ is able to cover the entire domain space with only $d_\star \ll d$ dimensions under the distribution $\fP$. If $\bphi$ is completely aligned with the domains themselves (i.e., we can easily separate the domains), we expect $d_\star \rightarrow 1$, and in the worst case, $d_\star = d$ (i.e., no information). Next, we present a standard assumption on the clustering approximation.
\begin{assumption}[Cluster Approximation, Informal]
\label{ass:cluster_sampling}
Let the optimal clustering (with infinite samples) of $\fP$ under $\bphi$ be given by $\bPsi_\star$, and let the optimal clustering (with $n\cdot N$ samples) from $\fP$ be given by $\widehat\bPsi_\star$. We assume that with high probability, $\lVert \bPsi_\star-\widehat\bPsi_\star\rVert_2 \leq \cO\left(\sqrt{\frac{1}{n\cdot N}}\right)$.
\end{assumption}
This is a standard assumption can be satisfied by most practical data distributions; for a thorough treatment of statistical clustering, see, e.g.,~\cite{Luxburg05towardsa, BenDavid2004AFF}. Armed with these assumptions, we present our primary generalization bound.

\begin{theorem}
\label{thm:main}
Let $\bphi$ and $\fP$ be such that Assumptions~\ref{ass:expressivity} and~\ref{ass:cluster_sampling} are true. Let $\cE_f =  \left| L(f, \bPsi) - \widehat L_N(f, \widehat\bPsi)\right|$ denote the generalization error for any $f\in\cF$. Then, with probability at least $1-\delta$,
{
\begin{align*}
    \sup_{f\in\cF}\cE_f = \cO\left(\left(\frac{1}{K}\right)^{\frac{1}{d_\star}} +  \sqrt{\frac{\log\left(\sfrac{KN}{\delta}\right)}{n}} + \sqrt{\frac{\log\left(\sfrac{nKN}{\delta}\right)}{N}}\right).
\end{align*}
}%
\end{theorem} 
\begin{remark}[Discussion]
\normalfont
The generalization bound above admits an identical dependency on the number of domains $N$ and points per domain $n$ as in prior work~\cite{dubey2021adaptive, blanchard2011generalizing}, and cannot be improved in general. We see an additional term $K^{-\frac{1}{d_\star}}$ which can be decomposed as follows. We see that as $K\rightarrow \infty$ (we select a larger clustering), the additional term goes to $0$. Its rate of decrease, however, depends on $d_\star$, i.e., the {\em effective dimensionality} of $\bphi$. If $\bphi$ contains ample information about $\fP$ (or $\fP$ is concentrated in $\bphi$), then we can expect $d_\star \ll d$ (it is at most $d$ by a covering bound). To the best of our knowledge, ours is the first analysis on {\em domain-adaptive} classification with kernel mean embeddings that considers the misspecification introduced by using approximate clustering solutions.
\end{remark}

A proof for Theorem~\ref{thm:main} can be found in the Appendix. In addition to the main result, we believe that the intermediary characterizations of aggregated clustering and expressivity can be of value beyond the domain generalization problem.
\begin{table*}[th!]
\centering
\small
\begin{tabular}{ccccccc}
\toprule
 \textbf{Algorithm} & VLCS~\cite{fang2013unbiased}        & PACS~\cite{li2017deeper}    & OffHome~\cite{venkateswara2017Deep}          & DNet~\cite{peng2019moment}           &  TerraInc~\cite{beery2018recognition}       &  Average               \\ \hline
\multicolumn{7}{c}{{\bf Model Selection:} Leave-One-Domain-Out Validation}\\
\hline
\multicolumn{7}{c}{Algorithms that {\em require} domain labels} \\
\hline
MLDG~\cite{li2017learning}     &76.8  $\pm$  0.4         &82.6 $\pm$  0.6         &67.7 $\pm$  0.3         &42.2 $\pm$  0.6         & 
46.0  $\pm$  0.4         &63.0  $\pm$  0.4         \\
CORAL~\cite{sun2016deep}    &77.3  $\pm$  0.3         &83.3 $\pm$  0.5         &68.6 $\pm$  0.1         &42.1 $\pm$  0.7         &
47.7  $\pm$  0.3         &63.8  $\pm$  0.3         \\
MMD~\cite{li2018domain}      &76.1  $\pm$  0.7         &83.1 $\pm$  0.7         &67.3 $\pm$  0.2         &38.7 $\pm$  0.5         &
45.8  $\pm$  0.6         &62.2  $\pm$  0.6         \\
C-DANN~\cite{li2018deep}   &73.8  $\pm$  1.1         &81.4 $\pm$  1.3         &64.2 $\pm$  0.5         &39.5 $\pm$  0.2         &
40.9  $\pm$  0.7         &60.1  $\pm$  0.9         \\ 
Mixup~\cite{wang2020heterogeneous}    &78.2  $\pm$  0.6         &83.9 $\pm$  0.8         &68.3 $\pm$  0.4         &40.2 $\pm$  0.4         &
46.2  $\pm$  0.5         &63.3  $\pm$  0.7         \\
DA-ERM~\cite{dubey2021adaptive}   &78.0  $\pm$  0.2         &84.1 $\pm$  0.5         & {\bf 67.9 $\pm$  0.4}         &{\bf 43.6 $\pm$  0.3}        &
47.3  $\pm$  0.5         &64.1  $\pm$  0.8         \\
\hline
\multicolumn{7}{c}{Algorithms that {\em do not require} domain labels} \\
\hline
ERM~\cite{gulrajani2020search}      &76.7  $\pm$   0.9         &83.2  $\pm$  0.7        &67.2 $\pm$  0.5         &41.1 $\pm$  0.8         &
46.2  $\pm$  0.3         &62.9  $\pm$  0.6         \\
IRM~\cite{arjovsky2019invariant}  &76.9  $\pm$  0.5         &82.8  $\pm$  0.5        &67.0 $\pm$  0.4         &35.7 $\pm$  1.9         & 
43.8  $\pm$  1.0         &61.2  $\pm$  0.8         \\
DRO~\cite{sagawa2019distributionally}     &77.3  $\pm$  0.2         &83.3 $\pm$  0.3         &66.8 $\pm$  0.1         &33.0 $\pm$  0.5         &
42.0  $\pm$  0.6         &60.4  $\pm$  0.4         \\
RSC~\cite{huang2020self} & 75.3 $\pm$ 0.5 & {\bf 87.7 $\pm$ 0.2} & 64.1 $\pm$ 0.4 & 41.2 $\pm$ 1.0 & 45.7 $\pm$ 0.3 &  62.8 $\pm$ 0.6\\
\hline
AdaClust($512 \rightarrow 1024$, 5) &{\bf 78.9 $\pm$ 0.6}   & 87.0 $\pm$  0.3  &67.7 $\pm$  0.5 & 43.3 $\pm$ 0.5 & {\bf 48.1  $\pm$  0.1} & {\bf 64.9  $\pm$  0.7}\\ 
\hline
\multicolumn{7}{c}{{\bf Model Selection:} Validation-Based Model Averaging (SWAD)~\cite{cha2021swad}}\\
\hline
 ERM   &78.8  $\pm$  0.1         &87.8$\pm$  0.2         &{\bf 69.8 $\pm$  0.1}         &46.0 $\pm$  0.1         &
50.2  $\pm$  0.3         &66.5  $\pm$  0.2         \\
 AdaClust($8 \rightarrow 520$, 5) &{\bf 79.6 $\pm$ 0.1}   & {\bf 89.2 $\pm$  0.4}  &69.4 $\pm$  0.2 & {\bf 46.7 $\pm$ 0.2} & {\bf 50.6  $\pm$  0.1} &{\bf 67.2  $\pm$  0.2}\\ 
\bottomrule
\end{tabular}
\caption{Benchmark Comparisons. All implementations are using {\sc DomainBed}~\cite{gulrajani2020search}. Experiments are repeated thrice with random seeds.}\label{tab:exp_small_scale}
\end{table*}

\section{Experiments}
Our experiments are performed using {\sc DomainBed}\cite{gulrajani2020search}, and the compared algorithms are run using hyperparameter ranges suggested therein. Our baseline experiments are done on a ResNet-50~\cite{he2016deep} neural network pre-trained on the ImageNet LSVRC12~\cite{deng2009imagenet} training set. 
\subsection{Experimental Setup}
\noindent{\bf Clustering}. We use the Facebook AI Similarlity Search (FAISS) \cite{JDH17} platform to perform clustering using the {\tt kmeans++} algorithm. To accelerate clustering, we quantize the feature space which introduces noise within the matching process. We set $K$ as an integer multiple of the number of classes in the problem (which depends on the dataset). We set the default clustering schedule $T_{\text{clust}} = \{1, 2, 4, 8, ...\}$, i.e., the first round of clustering occurs at epoch 1, then at epoch 2, then at epoch 4, and so on until training converges.

\noindent{\bf Training}. The feature extractor weights $\bomega$ we use are a ResNet-50 network pre-trained on ILSVRC12~\cite{deng2009imagenet} where we truncate the network after the {\tt pool5} layer. This layer is then concatenated with a $d_{\text{end}}-d_{\text{start}}$ dimensional pseudo-domain embedding, and passed through the fully-connected layer ($\bW$), with output dimension set to the number of classes. Note that $\bW$ is simply a fully-connected layer and not an MLP, as we did not find any notable performance improvements with the latter. We use weight decay with standard cross-entropy loss for training.

\subsection{Benchmark Comparisons}

\begin{figure*}[t]
\centering
\includegraphics[width=\linewidth]{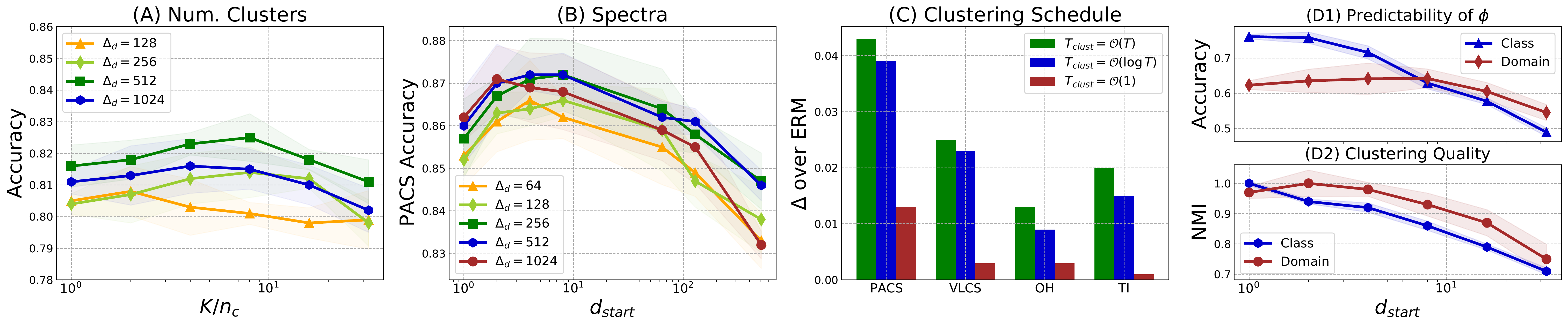}
\caption{Ablation Studies. (A) Comparison of average performance on PACS and VLCS when varying the number of clusters $K$ ($d_\text{start}$ is fixed); (B) Comparison on PACS when varying the spectrum starting index $d_{\text{start}}$ ($K$ is fixed); (C) Varying the clustering schedule $T_{\text{clust}}$; (D) Qualitative ablations on predictability of $\bphi$ and mutual information of clusters.}
\label{fig:qualitative_ablations}
\end{figure*}
We experiment on 5 datasets: VLCS~\cite{fang2013unbiased}, PACS~\cite{li2017deeper}, Office-Home~\cite{venkateswara2017Deep}, Domain-Net~\cite{peng2019moment} and Terra Incognita~\cite{beery2018recognition}). The variations within these datasets are on the nature of images (sketch vs. photographs) as well as synthetic (rotations and translations), providing a wide array of testing scenarios. Our default paradigm for model selection is the {\em leave-one-domain-out} cross validation method, where for a dataset with $N$ domains, we run $N$ experiments, wherein the $i^{th}$ round we leave the $i^{th}$ dataset out from training and only performing testing exclusively on it. Each domain is split $80-20$ at random into a training-validation split, following standard procedure outlined in {\sc DomainBed}\cite{gulrajani2020search}. We do a random search for the hyperparameters, which include neural network as well as clustering hyperparameters, and denote the specific hyperaparameters within the tables as AdaClust($d_{\text{start}} \rightarrow d_{\text{end}}, K/n_c$), i.e., spectrum range and clusters per class (where $n_c = |\cY|$). We report the full hyperparameter ranges in the Appendix.

{\bf Results} (Table~\ref{tab:exp_small_scale}). The algorithms we compare with are standard ERM (i.e., fine-tuning), {\em domain adaptive} classification~\cite{dubey2021adaptive}, and a suite of {\em domain-invariant} approaches pre-implemented in the {\sc DomainBed} suite. We observe that our algorithm, titled \textit{\textbf{Adaptive Clustering}}, performs competitively with all invariant approaches, as well as {\em domain-adaptive} classification, despite not utilizing any labels at all. Another line of research that provides competitive performance is based on {\em validation-based model averaging}, e.g., SWAD~\cite{cha2021swad}, which provides improvements orthogonal to our algorithm. This can be highlighted by running our algorithm with {\em validation-based model averaging} via SWAD, which is noted as \textit{\textbf{Adaptive Clustering}} (SWAD), and provides improvements over regular SWAD. 

\subsection{Large-Scale Comparisons}
In addition to the small-scale benchmarks from {\sc DomainBed}, we also provide comparisons on a real-world,  large-scale benchmark dataset to examine real-world performance. We use the {\bf GeoYFCC}~\cite{dubey2021adaptive} dataset, which is a real-world domain generalization benchmark with 1.1M total examples across 1.2K output categories and 62 total domains. We follow their outlined train/val/test splits and use traditional cross-validation on the heldout validation set to select hyperparameters. Note that given the size of the dataset, extensive hyperparameter tuning as employed by most algorithms is not possible, therefore it is a good benchmark to compare with {\em domain-invariant} approaches as well. We observe that our algorithm outperforms {\em domain invariant} approaches, matches the performance of the {\em domain-adaptive} {\bf DA-ERM}\cite{dubey2021adaptive} up to $0.1\%$, without domain labels. The results are summarized in Table~\ref{tab:yfcc}.
\subsection{Ablation Studies}
\label{sec:ablations}
\begin{table}
\small
\centering
\rowcolors{3}{gray!6}{white}
\begin{tabular}{ccc}
\toprule
\multirow{2}{*}{\textbf{Algorithm}} & Train & Test \\ 
& Top-1/5 & Top-1/5 \\
\hline
\multicolumn{3}{c}{Algorithms Using Domain Labels} \\ \hline
MMD & 25.4 / 50.9 & 21.8 / 46.2 \\
CORAL~\cite{sun2016deep} & 25.4 / 50.9 & 21.7 / 46.2 \\ 
DA-ERM~\cite{dubey2021adaptive} & 28.2 / 55.9 & {\bf 23.5 / 49.0} \\\hline
\multicolumn{3}{c}{Algorithms without Domain Labels} \\
\hline
ERM & 28.4 / 56.4 & 22.5 / 48.1 \\ 
\hline
AdaClust($8\rightarrow1032, 1$) & 28.4 / 56.2 & {\bf 23.4 / 48.9} \\
\bottomrule
\end{tabular}
\caption{Comparison on the Geo-YFCC~\cite{dubey2021adaptive} dataset. AdaClust matches the performance of DA-ERM without domain labels.}
\label{tab:yfcc}
\end{table}

\begin{table}
\footnotesize
\centering
\rowcolors{2}{gray!6}{white}
\begin{tabular}{cccccc}
\toprule
{\bf Algorithm} & VLCS        & PACS            & OH        & TI  & Avg.\\ 
\hline
ERM & 77.4 & 84.0 & 64.8 & 46.0 & 68.0 \\
AdaClust-Random & 76.5 & 83.1 & 63.6 & 45.2 &67.1\\
AdaClust-NoPCA & 77.5 & 84.2 & 65.3 & 46.9 & 68.5\\ \hline
AdaClust($8\rightarrow1032, 5$) & 78.2 & 86.2 & 65.2 & 48.1 & 69.4\\
\bottomrule
\end{tabular}
\caption{A comparison with various embedding approaches. AdaClust-Random refers to using random cluster embeddings (instead of nearest-neighbors), and AdaClust-NoPCA refers to clustering without using PCA. We see that {\em both} clustering and PCA are necessary for optimal adaptive clustering.}
\label{tab:nopca}
\end{table}

\subsubsection{Clustering Ablations}
\noindent{\bf Varying $K$} (Figure~\ref{fig:qualitative_ablations}A). We study the performance while changing the number of clusters $K$ while keeping $d_{\text{start}} = n_c$ and $d_{\text{end}} = \Delta_d + n_c$ fixed, where $n_c$ denotes the number of output classes for that dataset, and we try various $\Delta_d$. We compare performance for $K = n_c, 2n_c, 4n_c, 8n_c$ and $16n_c$ averaged on the PACS and VLCS datasets. We observe a fairly stable performance increase up to $K=8n_c$, after which it deteriorates, as we believe the recovered embeddings are too noisy when the number of clusters is large.

\noindent{\bf Varying $d_{\text{start}}$ and $d_{\text{end}}$} (Figure~\ref{fig:qualitative_ablations}B). We examine the impact of the indices $d_{\text{start}}$ and $d_{\text{end}}$ on generalization. For any $d_{\text{start}}, d_{\text{end}}$ pair, let $\Delta_d = d_{\text{end}} - d_{\text{start}}$ denote the spectrum width. We set $K=n_c$ and vary $d_{\text{start}} \in [0, 2, 4, 8, 64, 128, 512]$, $\Delta_d \in [64, 128, 256, 512, 1024]$ simultaneously on the PACS dataset. We observe the best performance at $d_{\text{end}} \in [256, 1024]$, and for any fixed $d_{\text{start}}$, we observe a small decrease as $d_{\text{end}}$ is increase beyond 512, potentially due to the addition of noisy features in later eigenvectors. We observe the best performance for all $d_{\text{start}}$ to be in the range $8-128$, which is in alignment with our hypothesis of introducing {\em class-independent} variance (see Section~\ref{sec:qualitative} for a qualitative analysis). 

\noindent{\bf Varying Clustering Schedule ($T_{\text{clust}}$)} (Figure~\ref{fig:qualitative_ablations}C). One expects that more rounds of clustering will improve performance at a higher clustering cost. To examine this, we compare 3 clustering schedules with $K=n_c$, $d_{\text{start}}=8$, $d_{\text{end}} = 520$ on the PACS dataset. We see no major differences in performance (validation accuracy within 0.2\%) between $T_{\text{clust}} = \cO(T)$ (cluster every epoch) and $T_{\text{clust}} = \cO(\log T)$ (doubling scheme with cost $\cO(\log T)$). However, if we cluster only a constant number of times (i.e., $T_{\text{clust}} = \cO(1)$) (where we perform clustering only once during the beginning, one halfway, and one at the end), we see very small improvements compared to baseline performances. The small deterioration due to a logarithmic $T_{\text{clust}}$ enables us to scale to large datasets without much difficulty.

\noindent{\bf Varying Clustering Algorithm}. We compare the performance due to different clustering subroutines, as it is known that different clustering algorithms often partition identical data differently~\cite{rokach2009survey}.  We examine this effect by testing out 4 different clustering algorithms on the PACS dataset, while keeping $K=n_c$, $d_{\text{start}} \in [0, 512, 1024]$, $d_{\text{end}} = d_{\text{start}} + 512$. We observe that {\tt kmeans++} clustering performs the best overall, whereas the other approaches (spectral clustering, agglomerative clustering and GMM) perform worse, and hence we select {\tt kmeans++}. This is also backed by the convergence guarantees obtained for {\tt kmeans++} with minimal assumptions on the data~\cite{choo2020k, wei2016constant}. The complete summary can be found in the Appendix.

\noindent{\bf Random Embeddings and Removing PCA} (Table~\ref{tab:nopca}). Finally, we compare two variants of the embeddings that are formed via other {\em unsupervised} approaches. We compare with {\em random} embeddings, i.e., select any domain embedding from the clustering at random (instead of the nearest), to examine the impact of additional cluster information. We also compare with a version where we do not run PCA, and cluster using the original features from $F_{\text{image}}$. We observe that in both cases, the performance is significantly worse than that with Adaptive Clustering, which indicates that {\em both} the clustering, and PCA are crucial for success. 
\begin{table}
\footnotesize
\centering
\begin{tabular}{ccccccc}
\toprule
{\bf Base} & \textbf{Alg.} & VLCS        & PACS            & OH        & TI  & Avg.\\ 
\hline
\multirow{2}{*}{ImageNet} & ERM                       & 77.4          & 84.0             & 64.8           & 46.0        & \multirow{2}{*}{{\color{ForestGreen} {\bf$\uparrow$1.3}}}                     \\
& AdaClust            & 78.2           & 86.2  & 65.2  & 48.1  &          \\
\midrule
\multirow{2}{*}{MS-Vision}&ERM                       & 76.7       & 88.8     & 70.8  & 47.6    & \multirow{2}{*}{{\color{ForestGreen} {\bf$\uparrow$2.1}}}                      \\
&AdaClust            & 80.4    & 90.5  & 72.4     & 49.1     &                     \\
\midrule
\multirow{2}{*}{FB-SWSL} &ERM                       & 78.5         & 88.2       & 69.2      & 48.0         & \multirow{2}{*}{{\color{ForestGreen} {\bf$\uparrow$2.4}}}                     \\
&AdaClust             & 79.7    & 91.3      & 71.4         & 49.5          &                  \\
\bottomrule
\end{tabular}
\caption{A comparison with vanilla ERM on different backbone networks. We see that the improvements with Adaptive Clustering improve as the expressivity of the backbone model increases.}
\label{tab:backbone}
\end{table}
\subsubsection{Backbone Ablations}

A central assumption in our approach is that we are pre-training from a large-scale model that contains the relevant information such that the domains are well-separated via $\bphi$. To examine this, we study the validation performance of our models on four different datasets: PACS, VLCS, OfficeHome and TerraIncognita, while we vary the backbone model used to train the network. The results are summarized in Table~\ref{tab:backbone} for three models: ResNet-50 on ILSVRC12 data~\cite{he2016deep}, the Microsoft Vision model~\cite{noauthor_microsoftvision:_nodate} that is trained on four distinct visual tasks, and the Facebook SWSL model~\cite{yalniz2019billion} that is weakly-supervised on 1B+ images. Since all models use the same architecture (ResNet-50), this comparison exclusively compares the large-scale nature of the problem. We observe an expected performance trend, with the Facebook SWSL model providing the best performance across datasets. More importantly, we see that compared with vanilla ERM, the relative improvement obtained by Adaptive Clustering increases with model size, highlighting that our algorithm is able to capture the relevant directions of variance residing in the pre-trained models. We set the hyperparameters as ($8\rightarrow1032, 5$).
\subsection{Qualitative Analysis}
\label{sec:qualitative}

We hypothesize that the variance in fine-tuned model feature space $\bphi$ follows the specific structure that the first few eigenvalues correspond to {\em class-specific} variance, followed by other structural variations, including {\em domain-specific} variance, followed eventually by noise inherited from the training process. We test this hypothesis with two sets of experiments. The first of these is measuring predictability of both the class and domain labels directly from the projected features $\bphi$. We consider $d_{\text{start}} \in [0, 1, 2, 3, 4, 8, 16, 32]$ and fix $K=N \times n_c$, $d_{\text{end}} = d_{\text{start}} + 256$, and train an MLP directly to predict the class and domain labels from $\bphi$, on the PACS dataset. We train a 1-layer MLP with SGD via leave-one-domain-out cross validation. Specifically, this looks at the alignment of features with all $256-$dimensional ``slices'' of the spectrum. The results of this ablation are summarized in Figure~\ref{fig:qualitative_ablations}(D1). We observe that the {\em domain} predictive power of $\bphi$ gradually increases as we remove the first few components, followed by a flat decline, whereas the {\em class} predictability of $\bphi$ decreases dramatically as the window is shifted.

Next, to compare the clustering quality, we examine the normalized mutual information (NMI)~\cite{rokach2005clustering} to examine the overlap between class partitions and domain partitions on the PACS dataset (the network is trained on all domains). We report the renormalized NMI (since $\bphi$ contains more information about the classes compared to domain labels) of the produced clusterings with both the class and domain labels. We observe that once again, as $d_{\text{start}}$ is shifted, the NMI with respect to domain partitions remains stable whereas the class NMI deteriorates rapdily, in line with our hypothesis.

\section{Discussion and Conclusion}

Domain generalization is an increasingly relevant problem in real-world settings that contain data restrictions motivated by privacy, such as the one studied in this paper. Our central contributions in this regard are as follows. First, we demonstrate that even when domain labels are not available, one can leverage large-scale pre-trained models to bootstrap for the domain generalization problem, and obtain competitive performance. Next, we extend the analysis of domain generalization via kernel mean embeddings to handle {\em approximate} embedding spaces such as the one presented in this paper. On their own, our contributions can provide interesting starting points for forays into other relevant problems, such as multi-task and multi-agent learning. 

Additionally, our contributions shed more light into the practical feasibility of {\em domain-invariant} learning: as discussed in~\cite{gulrajani2020search}, it is unclear whether explicitly modeling domain invariance outperforms naive ERM. Our research provides another argument in favor of naive ERM over modeling feature invariance, as we can see both experimentally and theoretically that whenever we have large pre-trained models to begin with, one can achieve competitive performance via ERM. Furthermore, as suggested by~\cite{dubey2021adaptive}, invariant approaches are difficult to scale to large-scale benchmarks, given their careful model selection requirements.

There are many follow-up directions that our work presents. First, we only explore clustering to partition the training data, whereas one can consider alternative approaches to compute the {\em pseudo-domain} embedding, including random projection hashing~\cite{rahimi2008random} or unsupervised random forests~\cite{pei2013unsupervised} to compute paritions faster. Alternatively, one can explore utilizing domain labels (whenever applicable) to accelerate the cluster discovery process. On the theoretical aspect, relaxing the expressivity assumption is a viable first step as well.
{\small
\bibliographystyle{ieee_fullname}
\bibliography{egbib}
}
\onecolumn
\appendix
\section{Details for Domain-Specific Filtering}
Consider an initial feature $\bomega_0$ (these can be features from a CNN pre-trained on ImageNet~\cite{deng2009imagenet}, for instance), which are gradually being fine-tuned on our training data $\widehat D_{\text{tr}}$. One can assume that the pre-final layer activations at any epoch $t$, denoted as $\bomega_t$ will gradually adapt from $\bomega_0$ so that they are discriminatory with respect to the classification task. Alternatively stated, this implies that the largest principal components of $\bPhi_t$ will gradually capture the {\em class-specific} variations within the data. Specifically, if the eigendecomposition of the feature $\bomega$ is given as follows:
\begin{equation*}
    \bPhi_t(\cdot) = \sum_{i=1}^d \lambda^t_i \cdot \be^t_i(\cdot),
\end{equation*}
\begin{equation*}
    \bV_t \bS_t \bV_t^\top = \frac{1}{n\cdot N -1}\bPhi_t(\bX)^\top\bPhi_t(\bX).
\end{equation*}
Where $\bS_t$ and $\bV_t$ are obtained by diagonalization, and $\bS_t$ is a diagonal matrix containing the eigenvalues of the scaled covariance matrix. Then, we compute the {\em end-truncated} eigenvectors $\bar\bV_t$ by only considering the middle $[d_{\text{start}}, d_{\text{end}}]$ columns of $\bV_t$ to create a matrix $\bar\bV_t \in \bbR^{d \times (d_{\text{end}} - d_{\text{start}})}$. The ``projected'' data points can then be obtained as:
\begin{equation*}
    \bar\bPhi_t(\bX) = \bPhi_t(\bX)\bar\bV_t.
\end{equation*}
Hence, any $\x$ can be projected by first computing its feature $\bPhi_t(\x)$ followed by a projection to get $\bphi_t(\x) = \bPhi_t(\x)\bar\bV_t$. The central idea with this step is to recover the ``Goldilocks zone'' of useful {\em domain-dependent} variance, such that when $d_{\text{start}}$ and $d_{\text{end}}$ are selected appropriately, the projected space is a good separator of different domains. Since we initialize $\bphi_0$ via a pre-trained network, trained originally on a large-scale dataset such as ImageNet~\cite{deng2009imagenet}, we hope that the diverse information present in these datasets allows us to obtain a useful embedding. As we see in our experiments, the algorithm is not too sensitive to the specific choice of $d_{\text{start}}$ and $d_{\text{end}}$, and is quite robust given a good starting model.

\section{Theoretical Guarantees}
In this section we provide some theoretical guarantees for our proposed algorithm. The theoretical results are within the framework of Blanchard \etal~\cite{blanchard2011generalizing}, where our key contributions are to analysing when the true kernel mean embeddings and obtained embeddings are mismatched. At a high level, our function class $\cF$ is defined over the space $\cX \times \cD$ (i.e., the joint input-domain space), however, we {\em embed} domains into $\bbR^d$ via a {\em mean embedding} $\bPhi : \cD \rightarrow \bbR^d$, and therefore, any function $f \in \cF$ is written as $f(\x, \bPsi_D)$, where $\bPsi_D$ denotes the {\em kernel mean embedding} of $D$.

\noindent{\bf Background}. We assume a {\em compact} input space $\cX$ and assume the outputs to lie in the space $\cY = [-1, 1]$. For any $L_\ell$-Lipschitz loss function $\ell : \bbR \times \cY \rightarrow \bbR_+$, then the empirical loss on any domain $D$ with $n$ samples is $\frac{1}{n_i}\sum_{(\x, y) \in D} \ell(f(\x, \bPsi_D), y)$. We can define the average training error over $N$ domains with $n$ samples as,
\begin{equation*}
    \widehat{L}_N(f, \bPhi) \triangleq \frac{1}{n\cdot N}\sum_{i \in [N]}\sum_{(\x, y) \in D_i} \ell(f(\x, \bPhi_{D_i}), y).
\end{equation*}
Similarly, our benchmark is to compare the above with the {\em expected} risk, i.e., the error obtained by $f$ in the limit of infinite samples.
\begin{equation*}
    L(f, \bPhi) \triangleq \underset{D\sim\fP}{\bbE}\left[\underset{(\x, y)\sim D}{\bbE}\left[\ell(f(\x, \bPsi_D), y\right]\right].
\end{equation*}
Following~\cite{blanchard2011generalizing}, the space $\cF$ we consider is defined by product kernels. Consider a P.S.D. kernel $\kappa$ over the product space $\cD_\cX \times \cX$\footnote{We assume there exist sets of probability distributions $\cD_\cX$ and $\cD_{\cY | \cX}$ such that for any sample $D \in \cD$ there exist samples $D_X \in \cD_\cX$ and $D_{Y|X} \in \cD_{\cY|\cX}$ such that $D = D_X \bullet D_{Y|X}$ (this characterization is applicable under suitable assumptions, see Section 3 of~\cite{blanchard2011generalizing}).} with associated RKHS $\cH_\kappa$. We select $f_\lambda$ such that
\begin{align*}
    f_\lambda  = \argmin_{f \in \cH_\kappa} \underbrace{\frac{1}{nN}\sum_{i=1}^N \sum_{j=1}^{n} \ell(f(\x_{ij},\widehat\bPsi), y_{ij})}_{\text{training error}} + \underbrace{\lambda\cdot \lVert f \rVert_{\cH_\kappa}^2}_{\text{regularization}}.
\end{align*}
Here, $\widehat\bPsi$ denotes the centroids obtained by pseudo-domain clustering, and $\kappa$ is a kernel on $\cD_\cX \times \cX$ defined as,
\begin{align*}
    \kappa((\x, \bPhi), (\x', \bPhi')) = f_\kappa(k_P(\bPhi, \bPhi'), k_X(\x, \x')).
    \label{eqn:product_kernels}
\end{align*}
Where $K_P$ and $K_X$ are kernels defined over $\fP_\cX$ and $\cX$ respectively, and $f_\kappa$ is Lipschitz in both arguments, with constants $L_P$ and $L_X$ with respect to the first and second argument respectively. Moreover, $K_P$ is defined with the use of yet another kernel $\fK$ that is a necessarily non-linear. For a feature vector $\bphi$, the kernel mean embedding $\bPsi_D$ of a domain $D$ is the image of $D$ in the RKHS of a distinct kernel $k'_X$ specified by $\bphi$, i.e.,
\begin{align*}
    D \rightarrow \Phi_D \triangleq \int_\cX k'_X(\x, \cdot) dD_X(\x) = \int_{\cX} \bphi(\x)dD_X(\x).
\end{align*}
We assume $k_X, k'_X$ and $\fK$ are bounded, i.e., $k_X(\cdot, \cdot) \leq B^2_k$, $k'_X(\cdot, \cdot) \leq B^2_{k'}$ and $\fK(\cdot, \cdot) \leq B^2_\fK$. 

\noindent{\bf Results}. Our key result is a uniform bound on the {\em excess risk} of using $K$ pseudo-domain centroids $\widehat\bPsi$ instead of the true kernel mean embeddings $\bPhi$, since computing the true kernel mean embeddings would require access to the domain labels. We introduce three constructs that are critical for our analysis. The first are the {\em optimal centroids} $\bPsi_\star = \{\bpsi^\star_k\}_{k=1}^K$ that are an optimal covering of the joint space $\cX_\bphi = \{\bphi(\x) | \x\in\cX\}$ in expectation, i.e., 
\begin{align*}
  \bPsi_\star = \argmin_{\bTheta \in \cB_d(1)^K} \underset{D\sim\fP}{\bbE}\left[\underset{\x\sim D}{\bbE} \left[\min_{k\in[K]} \lVert \bphi(\x) - \bTheta_{[k]} \rVert _2\right]\right].
\end{align*}
We denote the R.H.S. cost as $\cC(\bTheta)$ for brevity. Note that since $\cX_\bphi \subseteq \cB_d(1)$ (as $\lVert \bphi(\cdot)\rVert_2 \leq 1$), $\cC(\bPsi_\star) \leq \frac{3}{K^{\sfrac{1}{d}}}$ even in the worst case, by a simple covering argument. In a nutshell, $\bPsi_\star$ denotes the best possible {\em aggregated} clustering. Next, we define the {\em optimal domain-wise centroids}, $\widetilde\bPsi_\star = \{\tilde\bpsi^\star_k\}_{k=1}^K$ that denote the optimal covering of the data space partitioned by domains, i.e,
\begin{align*}
  \widetilde\bPsi_\star = \argmin_{\bTheta \in\cB_d(1)^K} \underset{D\sim\fP}{\bbE}\left[ \min_{k\in[K]}\left\lVert \underset{\x\sim D}{\bbE} \left[\bphi(\x)\right] - \bTheta_{[k]} \right\rVert_2\right].
\end{align*}
We denote the R.H.S. cost as $\widetilde\cC(\bTheta)$ for brevity. Note that these centroids are different from the ones earlier, since these consider the centroids that are closest to the domain-wise embeddings for each $D\in\cD$. Nevertheless, since $\cD_\bphi \subseteq \cB_d(1)$ (as $\lVert \bphi(\cdot)\rVert_2 \leq 1$), $\widetilde\cC(\widetilde\bPsi_\star) \leq \frac{3}{K^{\sfrac{1}{d}}}$ even in the worst case, by a simple covering argument. Finally, we define the $K$ {\em optimal sample centroids} $\widehat\bPsi_\star = \{\widehat\bpsi^\star_k\}_{k=1}^K$ for any set $\widehat\cX = \{\x_{ij}\}_{i, j}^{N, n}$ that denote the optimal covering of the training data, i.e.,
\begin{align*}
  \widehat\bPsi_\star = \argmin_{\bTheta \in\cB_d(1)^K} \sum_{i=1}^N\sum_{j=1}^n\min_{k\in[K]} \lVert \bphi(\x_{ij}) - \bTheta_{[k]} \rVert _2.
\end{align*}
Note that all three of $\bPsi_\star, \widetilde\bPsi_\star$ and $\widehat\bPsi_\star$ are {\em independent} of the clustering algorithms, and {\em only} depend on the data distribution $\fP$ and feature $\bphi$. We now define some terms based on this notation.
\begin{definition}[Bad Neighbors]
We define any pair of points as {\bf bad neighbors} if they belong to the same cluster in $\bPsi_\star$ but not in $\widetilde\bPsi_\star$ or vice-versa. Specifically, if, for any point $(\x, D)$ we denote the nearest centroid in $\bPsi_\star$ as $\bpsi^\star_\x$ and nearest centroid in $\widetilde\bPsi_\star$ as $\tilde\bpsi^\star_D$, then any pair of points $(\x, D)$ and $(\x', D')$ are bad neighbors if either of the following are true:
\begin{align*}
    \left\{\bpsi^\star_\x = \bpsi^\star_{\x'} \text{ and } \bpsi^\star_D \neq \bpsi^\star_{D'}\right\} \text{ or } \left\{\bpsi^\star_\x \neq \bpsi^\star_{\x'} \text{ and } \bpsi^\star_D = \bpsi^\star_{D'}\right\}.
\end{align*}
\end{definition}
The above definition identifies points that lie within different domain clusters (or data clusters) but are in the same data cluster (or domain cluster), and we use this definiton to rigorously define how expressive the feature embedding $\bphi$ is in separating the domains.
\begin{assumption}[$d_\star-$Expressivity]
\label{ass:expressivity}
Let $\bbP_0$ denote the probability that two i.i.d. sampled points $(\x, D)$ and $(\x', D')$ from $\fP$ are bad neighbors. We then assume that the feature space $\bphi$ is such that there exists a constant $d_\star \ll d$ that satisfies, 
\begin{align*}
    \cC(\bPsi_\star) = \cO\left(\frac{1}{K^{\frac{1}{d_\star}}}\right) \text{ and } \bbP_0 = \cO\left(\frac{\sigma_\fP}{K^{\frac{1}{d_\star}}}\right),
\end{align*}
where $\sigma_\fP = \bbE_{D\sim\fP}\bbE_{\x\sim D}[\lVert \bphi(\x)\rVert_2^2]^{1/2}$ denotes the effective standard deviation of $\bphi$.
\end{assumption}
\begin{remark}[Expressivity]
\normalfont
The expressivity condition is how we formalize the notion of {\em domain-dependent} variance present within the feature space $\bphi$. It implies that $\bphi$ is able to cover the entire domain space with only $d_\star \ll d$ dimensions under the distribution $\fP$. If $\bphi$ is completely aligned with the domains themselves (i.e., we can easily separate the domains), we expect $d_\star \rightarrow 1$, and in the worst case, $d_\star = d$ (i.e., no information).
\end{remark}
Next, we present a standard assumption on the clustering approximation.
\begin{assumption}[Cluster Sampling Approximation]
\label{ass:cluster_sampling}
We assume that there exist absolute constants $C$ such that with probability at least $1-\delta$, $\delta \in (0, 1)$,
\begin{equation*}
    \max_{k, k' \in [K]} \left\lVert \bPsi_{\star, [k]} - \widehat\bPsi_{\star, [k']} \right\rVert_2 \leq C\cdot\sqrt{\frac{\log\left(\tfrac{nKN}{\delta}\right)}{nN}}.
\end{equation*}
\end{assumption}
This is a standard assumption can be satisfied by most practical data distributions; for a thorough treatment of statistical clustering, see, e.g.,~\cite{Luxburg05towardsa, BenDavid2004AFF}. Armed with these assumptions, we present our primary generalization bound.

\begin{theorem}
Let $\bphi$ and $\fP$ be such that $\cC(\bPsi_\star) = o\left(K^{-\frac{1}{d_\star}}\right)$ for $d_\star \leq d$, and Assumptions~\ref{ass:expressivity} and~\ref{ass:cluster_sampling} are true. Let $\cE_f =  \left| L(f, \bPhi) - \widehat L_N(f, \widehat\bPsi)\right|$ denote the generalization error for any $f\in\cF = \cB_\kappa(1)$. Then, with probability at least $1-\delta$,
{
\begin{align*}
    \sup_{f\in\cF}\cE_f = \cO\left(K^{\frac{-1}{pd_\star}} + \Delta + \sqrt{\tfrac{\log\left(\sfrac{KN}{\delta}\right)}{n}} + \sqrt{\tfrac{\log\left(\sfrac{nKN}{\delta}\right)}{N}}\right).
\end{align*}
}%
\end{theorem} 

\begin{remark}[Discussion]
\normalfont
The generalization bound above admits an identical dependency on the number of domains $N$ and points per domain $n$ as in prior work~\cite{dubey2021adaptive, blanchard2011generalizing}, and cannot be improved in general. We see an additional term $K^{-\frac{1}{d_\star}}$ which can be decomposed as follows. We see that as $K\rightarrow \infty$ (we select a larger clustering), the additional term goes to $0$. Its rate of decrease, however, depends on $d_\star$, i.e., the {\em effective dimensionality} of $\bphi$. If $\bphi$ contains ample information about $\fP$ (or $\fP$ is concentrated in $\bphi$), then we can expect $d_\star \ll d$ (it is at most $d$ by a covering bound). To the best of our knowledge, ours is the first analysis on {\em domain-adaptive} classification with kernel mean embeddings that considers the misspecification introduced by using approximate clustering solutions.
\end{remark}
\section{Proof of Theorem~\ref{thm:main}}
We begin by providing some assistive lemmas.
\begin{lemma}
\label{lem:clustering_bound}
For any set of $K$ centroids $\bPsi = \{\bpsi_1, ..., \bpsi_K\}, \bPsi \in \bbR^{dK}$ such that $\forall k \in [K]$, $\lVert \bpsi_k \rVert_2 \leq 1$, and feature $\bphi(\cdot) \in \bbR^d$ with $\lVert \bphi(\cdot) \rVert_2 \leq 1$, let the {\em expected} and {\em empirical} clustering cost be given as,
\begin{align*}
    \cC(\bPsi) =  \bbE_{D\sim\fP}\left[\bbE_{\x\sim D}[c(\x, \bPsi)]\right], \widehat\cC(\bPsi) = \frac{1}{n \cdot N}\sum_{i=1}^N\sum_{j=1}^n c(\x_{ij}, \bPsi),
\end{align*}
where $c(\x, \bPsi) = \min_{k\in[K]} \lVert \bphi(\x) -  \bPsi_k\rVert_2$. Then, we have with probability at least $1-\delta$,
\begin{align*}
    \left|\cC(\bPsi)- \widehat\cC(\bPsi)\right| \leq 2\sqrt{\frac{\log(2N/\delta)}{n}} + 2\sqrt{\frac{\log(1/\delta)}{N}}.
\end{align*}
\end{lemma}
\begin{proof}
Observe that for any $\bphi(\x) \in \cB_d(1)$ and fixed $\bPsi \in \cB_d(1)^K$, $c(\x, \bPsi) = \min_{k\in[K]} \lVert \bphi(\x) -  \bPsi_k\rVert_2$ is bounded in $[0, 2]$, and hence $c(\x, \bPsi)$ is sub-Gaussian with variance proxy at most $2$. Since $\x$ are sampled i.i.d. from $D$, we have, by Hoeffding's inequality, for any domain $D_i$ with probability at least $1-\delta'$,
\begin{align*}
    \bbE_{\x\sim D_i}\left[c(\x, \bPsi)\right] &\leq \frac{1}{n}\sum_{j=1}^n c(\x_{ij}, \bPsi) + 2\sqrt{\frac{\log(1/\delta')}{n}}.
\end{align*}
Furthermore, for any domain $D$, the variable $\bbE_{\x\sim D}[c(\x, \bPsi)]$ is also 2 sub-Gaussian (by the boundedness of $c$), and therefore, we have, with probability at least $1-\delta''$,
\begin{align*}
    \bbE_{D\sim\fP}\left[\bbE_{\x\sim D}[c(\x, \bPsi)]\right] &\leq \frac{1}{N}\sum_{i=1}^N \bbE_{\x\sim D_i}[c(\x, \bPsi)]+ 2\sqrt{\frac{\log(1/\delta'')}{N}}.
\end{align*}
We can bound the first term in the R.H.S. by taking a union bound over the first result and setting $\delta' = \delta/2N$, $\delta'' = \delta/2$. We have, with probability at least $1-\delta$,
\begin{align*}
    \bbE_{D\sim\fP}\left[\bbE_{\x\sim D}[c(\x, \bPsi)]\right] &\leq \frac{1}{n \cdot N}\sum_{i=1}^N\sum_{j=1}^n c(\x_{ij}, \bPsi) + 2\sqrt{\frac{\log(2N/\delta)}{n}} + 2\sqrt{\frac{\log(1/\delta)}{N}}.
\end{align*}
Substituting shorthand notations provides us the result. We can derive the opposite direction in an identical manner.
\end{proof}

\begin{lemma}
\label{lem:kme_bound}
For any point $\x\in\cX$, let $\bpsi_\x = \argmin_{k\in[K]}\lVert\bphi(\x) - \bpsi^\star_k\rVert_2$ denote the optimal cluster it is closest to. Then we have that,
\begin{align*}
    \bbE_{D\sim\fP}\left[\bbE_{\x\sim D}\left[ \left\lVert \bpsi_\x - \bbE_{\x\sim D}[\bphi(\x)]\right\rVert_2\right]\right] &= \cO\left(\frac{\sigma_\fP}{K^{\frac{1}{d_\star}}}\right).
\end{align*}
\end{lemma}
\begin{proof}
For any domain $D$, let $\tilde\bpsi_D$ denote the centroid nearest to $D$ from $\widetilde\bPsi$. We have that,
\begin{align*}
    \bbE_{D\sim\fP}\left[\bbE_{\x\sim D}\left[ \left\lVert \bpsi_\x - \bbE_{\x\sim D}[\bphi(\x)]\right\rVert_2\right]\right] &= \bbE_{D\sim\fP}\left[\bbE_{\x\sim D}\left[ \left\lVert \bpsi_\x -\tilde\bpsi_D + \tilde\bpsi_D - \bbE_{\x\sim D}[\bphi(\x)]\right\rVert_2\right]\right] \\
    &\leq \bbE_{D\sim\fP}\left[\bbE_{\x\sim D}\left[ \left\lVert \bpsi_\x -\tilde\bpsi_D\right\rVert_2\right]\right] + \bbE_{D\sim\fP}\left[\bbE_{\x\sim D}\left[ \left\lVert\tilde\bpsi_D - \bbE_{\x\sim D}[\bphi(\x)]\right\rVert_2\right]\right] \\
    &\leq \bbE_{D\sim\fP}\left[\bbE_{\x\sim D}\left[ \left\lVert \bpsi_\x -\tilde\bpsi_D\right\rVert_2\right]\right] + \frac{C}{K^{\frac{1}{d_\star}}}.
\end{align*}
The last inequality follows from Assumption~\ref{ass:expressivity}. Now, to bound the first term, for any centroid $\bpsi$, let $\cX_{\bpsi}$ denote the partition of $\cX$ it covers (and similarly for $\cD$). Observe that we can write the centroids $\bpsi_\x = \int_{\y\in\cX_{\bpsi_\x}} \x p(\y)$ and $\tilde\bpsi_D = \int_{D'\in\cD_{\tilde\bpsi_D}}\bPhi_{D'} p(D')$, as they are the average features within those regions. Substituting this gives us, for any $\x, D \in \cX \times \cD$, the inner term in the summation,
\begin{align*}
     \left\lVert \bpsi_\x -\tilde\bpsi_D\right\rVert_2 &= \left\lVert \int_{D'\in\cD}\int_{\x'\in\cX} \bphi(\x')\left[\bone\{\x'\in\cX_{\bpsi_\x}\} - \bone\{D'\in\cD_{\bpsi_D}\}\right]p(\x', D')\right\rVert_2 \\
     &\leq \int_{D'\in\cD}\int_{\x'\in\cX} \left\lVert \bphi(\x')\left[\bone\{\x'\in\cX_{\bpsi_\x}\} - \bone\{D'\in\cD_{\bpsi_D}\}\right]p(\x', D')\right\rVert_2 &\tag{Jensen} \\
     &= \int_{D'\in\cD}\int_{\x'\in\cX} \left\lVert \bphi(\x')\right\rVert_2\cdot\left|\bone\{\x'\in\cX_{\bpsi_\x}\} - \bone\{D'\in\cD_{\bpsi_D}\}\right|p(\x', D') \\
     &\leq \left(\int_{D'\in\cD}\int_{\x'\in\cX} \left\lVert \bphi(\x')\right\rVert_2p(\x', D')\right)\cdot\left(\int_{D'\in\cD}\int_{\x'\in\cX}\left|\bone\{\x'\in\cX_{\bpsi_\x}\} - \bone\{D'\in\cD_{\bpsi_D}\}\right|p(\x', D')\right) &\tag{Cauchy-Schwarz} \\
     &= \sigma_\fP\cdot\left(\int_{D'\in\cD}\int_{\x'\in\cX}\bone\{\x'\in\cX_{\bpsi_\x} \oplus D'\in\cD_{\bpsi_D}\}p(\x', D')\right) \\
     &= \sigma_\fP\cdot\bbP_{\x', D'}\{\x'\in\cX_{\bpsi_\x} \oplus D'\in\cD_{\bpsi_D}\}.
\end{align*}
Replacing this in the expectation, we have that, for some absolute constant $C'$, since $\x$ and $\x'$ (resp. $D$ and $D'$) are independent,
\begin{align*}
     \bbE_{D\sim\fP}\left[\bbE_{\x\sim D}\left[\sigma_\fP\cdot\bbP_{\x', D'}\{\x'\in\cX_{\bpsi_\x} \oplus D'\in\cD_{\bpsi_D}\}\right]\right] &= \sigma_\fP\cdot\bbP_0 \leq \frac{\sigma_\fP\cdot C'}{K^{\frac{1}{d_\star}}}.
\end{align*}
Putting it together gives us the final result.
\end{proof}
We are now ready to prove Theorem~\ref{thm:main}.
\begin{proof}
We begin by decomposing the LHS. 
\begin{align*}
    &\sup_{f \in \cB_\kappa(R)} \left| L(f, \bPhi) - \widehat{L}_N(f, \widehat\bPsi) \right| \\
    &= \sup_{f \in \cB_\kappa(R)} \left| L(f, \bPhi) - \widehat{L}_N(f, \bPsi) + \widehat{L}_N(f, \bPsi) - \widehat{L}_N(f, \widehat\bPsi_\star) + \widehat{L}_N(f, \widehat\bPsi_\star) - \widehat{L}_N(f, \widehat\bPsi) \right| \\
    &\leq \underbrace{\sup_{f \in \cB_\kappa(R)} \left| L(f, \bPhi) - \widehat{L}_N(f, \bPsi) \right|}_{\textcircled{1}} + \underbrace{\sup_{f \in \cB_\kappa(R)} \left| \widehat{L}_N(f, \bPsi) - \widehat{L}_N(f, \widehat\bPsi_\star) \right|}_{\textcircled{2}} + \underbrace{\sup_{f \in \cB_\kappa(R)} \left| \widehat{L}_N(f, \widehat\bPsi_\star) - \widehat{L}_N(f, \widehat\bPsi) \right|}_{\textcircled{3}}
\end{align*}
We first bound the term $\textcircled{3}$. For any $K$ clustering, let the set of points assigned to cluster $k$ be given by $n_k$. Observe that, since the loss function $L$ is Lipschitz in $f$, we have,
\begin{align*}
    \underset{f \in \cB_\kappa(R)}{\sup} |\hat{L}_N(f, \widehat\bPsi_\star) - \hat{L}_N(f, \widehat\bPsi)| & \leq \frac{L_\ell}{nN} \cdot \underset{f \in \cB_\kappa(R)}{\sup} \left| \sum_{i=1}^N\sum_{j=1}^n{f(\x_{ij}, \widehat\bpsi^\star_\x) - f(\x_{ij}, \widehat\bpsi_\x) }\right| \\
    & \leq \frac{L_\ell}{nN}  \sum_{i=1}^K\sum_{j=1}^{n_k}\underset{f \in \cB_\kappa(R)}{\sup} \left|f(\x_{ij}, \widehat\bpsi^\star_\x) - f(\x_{ij}, \widehat\bpsi_\x) \right|.
\end{align*}
Now, by the reproducing property of $\kappa$ (and the corresponding RKHS), we have that for any $\x \in \cX$ that has an optimal empirical centroid $\widehat\bpsi^\star_\x$, and {\em estimated} empirical centroid $\widehat\bpsi_\x$,
\begin{align*}
    \underset{f \in \cB_\kappa(R)}{\sup} \left|f(\x, \widehat\bpsi^\star_\x) - f(\x, \widehat\bpsi_\x) \right| &\leq \lVert f \rVert_{\kappa} \sup \left|f_\kappa(k_P(\widehat\bpsi^\star_\x, \cdot), k_X(\x, \cdot))- f_\kappa(k_P(\widehat\bpsi_\x, \cdot), k_X(\x, \cdot))\right| \\
    &\leq R \cdot\sup \left|f_\kappa(k_P(\widehat\bpsi^\star_\x, \cdot), k_X(\x, \cdot))- f_\kappa(k_P(\widehat\bpsi_\x, \cdot), k_X(\x, \cdot))\right| \tag{Since $\lVert f\rVert_\kappa \leq R$}\\
    &\leq RL_P \cdot\sup \left| \fK(\widehat\bpsi^\star_\x, \cdot) - \fK(\widehat\bpsi_\x, \cdot) \right| \tag{Since $f_\kappa$ is Lipschitz} \\
    &\leq RL_P \cdot\sup \left\lVert \Phi_\fK(\widehat\bpsi^\star_\x) - \Phi_\fK(\widehat\bpsi_\x)\right\rVert \tag{Triangle inequality}\\
    &\leq RL_P \cdot\sup \left\lVert \widehat\bpsi^\star_\x - \widehat\bpsi_\x\right\rVert_\infty \tag{$1$-H\"older assumption}.
\end{align*}
By replacing this result above, we have that, with probability at least $1-\delta$,
\begin{align*}
    \frac{L_\ell}{nN}  \sum_{i=1}^K\sum_{j=1}^{n_k}\underset{f \in \cB_\kappa(R)}{\sup} \left|f(\x_{ij}, \widehat\bpsi^\star_\x) - f(\x_{ij}, \widehat\bpsi_\x) \right| &\leq \frac{L_\ell}{nN}RL_P  \sum_{i=1}^K\sum_{j=1}^{n_k} \left\lVert \widehat\bpsi^\star_\x - \widehat\bpsi_\x\right\rVert_\infty \\
    &= \frac{L_\ell RL_P}{nN}  \sum_{i=1}^Kn_k\cdot \left\lVert \widehat\bpsi^\star_\x - \frac{1}{n_k}\sum_{j=1}^{n_k}\bphi(\x_{ij})\right\rVert_\infty \\
    &\leq \frac{L_\ell RL_P}{nN}  \sum_{i=1}^K\sum_{j=1}^{n_k}\left\lVert \widehat\bpsi^\star_\x - \bphi(\x_{ij})\right\rVert_2 \tag{$\lVert \cdot \rVert_\infty \leq \lVert \cdot \rVert_2$}\\
    &\leq L_\ell RL_P \cdot \widehat\cC(\widehat\bPsi_\star) \tag{H\"older's inequality} \\
    &\leq L_\ell RL_P \cdot\widehat\cC(\bPsi) \tag{$\widehat\bPsi_\star$ minimizes $\widehat\cC$} \\
    &\leq L_\ell RL_P \cdot\left(\cC(\bPsi) + 2\sqrt{\frac{\log(2N/\delta)}{n}} + 2\sqrt{\frac{\log(1/\delta)}{N}} \right) \tag{Lemma~\ref{lem:clustering_bound}} \\
    &\leq L_\ell RL_P \cdot\left(\frac{C''}{K^{1/d_\star}} + 2\sqrt{\frac{\log(2N/\delta)}{n}} + 2\sqrt{\frac{\log(1/\delta)}{N}} \right) \tag{Assumption~\ref{ass:expressivity}}.
\end{align*}
Now, we bound term $\textcircled{2}$. By a similar decomposition as earlier, we have that with probability at least $1-\delta$,
\begin{align*}
    \sup_{f \in \cB_\kappa(R)} \left| \widehat{L}_N(f, \bPsi) - \widehat{L}_N(f, \widehat\bPsi_\star) \right| &\leq \frac{L_\ell}{nN}RL_P  \sum_{i=1}^K\sum_{j=1}^{n_k} \left\lVert \widehat\bpsi^\star_\x - \bpsi_\x\right\rVert_\infty \\
    &\leq L_\ell L_P R \cdot\left(C\sqrt{\frac{\log\left(\tfrac{nKN}{\delta}\right)}{nN}}\right). \tag{Assumption~\ref{ass:cluster_sampling}}
\end{align*}
Finally, we arrive to bounding term $\textcircled{1}$. This term can further be decomposed as follows.
\begin{align*}
    \sup_{f \in \cB_\kappa(R)} \left| L(f, \bPhi) - \widehat{L}_N(f, \bPsi) \right| &\leq \underbrace{\sup_{f \in \cB_\kappa(R)} \left| L(f, \bPhi) - L(f, \bPsi) \right|}_{\textcircled{A}} + \underbrace{\sup_{f \in \cB_\kappa(R)} \left| L(f, \bPsi) - \widehat{L}_N(f, \bPsi) \right|}_{\textcircled{B}}.
\end{align*}
Here, the first term $\textcircled{A}$ measures the difference in expected error when $f$ uses $\bPsi$ instead of the true kernel mean embeddings $\bPhi$. We see that this can be bound by Assumption~\ref{ass:expressivity}.
\begin{align*}
    \sup_{f \in \cB_\kappa(R)} \left| L(f, \bPhi) - L(f, \bPsi) \right| &\leq \sup_{f\in\cB_\kappa(R)}\bbE_{D\sim\fP}\left[\bbE_{\x\sim D}\left[\left|L(\x, \bPsi_D) - L(\x, \bpsi_\x)\right|\right]\right] \\
    &\leq L_\ell\cdot\bbE_{D\sim\fP}\left[\bbE_{\x\sim D}\left[\sup_{f\in\cB_\kappa(R)}\left|\ell(\x, \bPsi_D) - \ell(\x, \bpsi_\x)\right|\right]\right] \tag{$\ell$ is $L_\ell$-Lipschitz} \\
    &\leq R \cdot L_\ell\cdot\bbE_{D\sim\fP}\left[\bbE_{\x\sim D}\left[\sup \left|f_\kappa(k_P(\bpsi_\x, \cdot), k_X(\x, \cdot))- f_\kappa(k_P(\bPsi_D, \cdot), k_X(\x, \cdot))\right|\right]\right] \tag{Since $\lVert f\rVert_\kappa \leq R$}\\
    &\leq RL_P L_\ell\cdot\bbE_{D\sim\fP}\left[\sup \left| \fK(\widehat\bpsi_\x, \cdot) - \fK(\bPsi_D, \cdot) \right|\right] \tag{Since $f_\kappa$ is Lipschitz} \\
    &\leq RL_PL_\ell\cdot\bbE_{D\sim\fP}\left[\bbE_{\x\sim D}\left[ \sup \left\lVert \Phi_\fK(\bpsi_\x) - \Phi_\fK(\bPsi_D)\right\rVert\right]\right] \tag{Triangle inequality}\\
    &\leq RL_PL_\ell\cdot\bbE_{D\sim\fP}\left[\bbE_{\x\sim D}\left[ \left\lVert \bpsi_\x - \bbE_{\x\sim D}[\bphi(\x)]\right\rVert_\infty\right]\right] \tag{$1$-H\"older assumption}.\\
    &\leq \frac{C'\sigma_\fP RL_PL_\ell}{K^{\frac{1}{d_\star}}}. &\tag{Lemma~\ref{lem:kme_bound}}
\end{align*}
Term $\textcircled{B}$ can be bound identically to Section 3.2 of Blanchard \etal~\cite{blanchard2011generalizing} (appendix). Putting everything together, we have the final result.
\end{proof}

\section{DomainBed Hyperparameters}
We follow Gulrajani and Lopez-Paz~\cite{gulrajani2020search} for hyperparameters. These values are summarized in Table \ref{table:hyperparameters}.

\begin{table}[H]
    \begin{center}
    { 
    \begin{tabular}{llll}
        \toprule
        \textbf{Condition} & \textbf{Parameter} & \textbf{Default value} & \textbf{Random distribution}\\
        \midrule
        \multirow{3}{*}{Basic hyperparameters}       & learning rate & 0.00005 & $10^{\text{Uniform}(-5, -3.5)}$\\
                                      & batch size    & 32   & $2^{\text{Uniform}(3, 5.5)}$\\
                                      & weight decay & 0    & $10^{\text{Uniform}(-6, -2)}$\\
        \midrule
        \multirow{8}{1.5cm}{C-DANN} & lambda                 & 1.0    & $10^{\text{Uniform}(-2, 2)}$\\
                                      & generator learning rate & 0.00005 & $10^{\text{Uniform}(-5, -3.5)}$\\
                                      & generator weight decay & 0    & $10^{\text{Uniform}(-6, -2)}$\\
                                      & discriminator learning rate & 0.00005 & $10^{\text{Uniform}(-5, -3.5)}$\\
         & discriminator weight decay & 0    & $10^{\text{Uniform}(-6, -2)}$\\
         & discriminator steps        & 1    & $2^{\text{Uniform}(0, 3)}$\\
         & gradient penalty           & 0    & $10^{\text{Uniform}(-2, 1)}$\\
         & adam $\beta_1$             & 0.5    & $\text{RandomChoice}([0, 0.5])$\\
        \midrule
        \multirow{2}{*}{IRM}          & lambda  & 100    & $10^{\text{Uniform}(-1, 5)}$\\
                                      & iterations of penalty annealing & 500 & $10^{\text{Uniform}(0, 4)}$\\
        \midrule
        Mixup                         & alpha & 0.2 & $10^{\text{Uniform}(0, 4)}$\\
        \midrule
        DRO                           & eta   & 0.01 & $10^{\text{Uniform}(-1, 1)}$\\
        \midrule
        MMD                           & gamma & 1 & $10^{\text{Uniform}(-1, 1)}$\\
        \midrule
        MLDG           & beta & 1 & $10^{\text{Uniform}(-1, 1)}$\\
        \midrule
        all          & dropout & 0    & $\text{RandomChoice}([0, 0.1, 0.5])$\\
        \bottomrule
    \end{tabular}
    }
    \end{center}
    \caption{{\sc DomainBed} hyperparameters.} 
    \label{table:hyperparameters}
\end{table}
\section{Geo-YFCC Hyperparameters}
We do a grid search on number of epochs in $(3, 6, 12, 25)$ and fixed best performance at 6 epochs. We have a batch size of 80 for all methods (AdaClust, ERM, DA-ERM, MMD, CORAL). We fix the learning rate as $0.04$ and weight-decay of $1e-5$ over a system of 64 GPUs. We tune the value of the loss weight in the range $(0, 1.5)$ for MMD and CORAL. The reported results are with the loss weight ($\gamma = 1$).
\section{Varying the Clustering Algorithm}

The ablations on varying the clustering algorithm are summarized in Table~\ref{tab:clustering}.
\begin{table}
\footnotesize
\centering
\rowcolors{2}{gray!6}{white}
\begin{tabular}{cccccc}
\toprule
{\bf Algorithm} & VLCS        & PACS            & OH        & TI  & Avg.\\ 
\hline
ERM & 77.4 & 84.0 & 64.8 & 46.0 & 68.0 \\
AdaClust-GMM & 76.8 & 84.9 & 64.8 & 47.0 &68.3\\
AdaClust-Agglomerative & 77.8 & 85.7 & 65.9 & 46.3 & 68.9\\ \hline
AdaClust(k-means++) & 78.2 & 86.2 & 65.2 & 48.1 & 69.4\\
\bottomrule
\end{tabular}
\caption{A comparison with various clustering approaches.}
\label{tab:clustering}
\end{table}
\end{document}